\documentclass[10pt,twocolumn,letterpaper]{article}

\usepackage[pagenumbers]{wacv} 

\usepackage{amsmath}
\usepackage{algorithm}
\usepackage{algpseudocode}

\usepackage{makecell}

\usepackage{lipsum}
\usepackage{balance}

\definecolor{wacvblue}{rgb}{0.21,0.49,0.74}
\usepackage[pagebackref,breaklinks,colorlinks,allcolors=wacvblue]{hyperref}

\def\wacvPaperID{235} 
\def\confName{WACV}
\def\confYear{2026}

\title{P3P Made Easy}

\author{Seong Hun Lee, \ \ Patrick Vandewalle 
\\
EAVISE, KU Leuven\\
Jan Pieter De Nayerlaan 5,
Sint-Katelijne-Waver, Belgium\\
{\tt\small \{seonghun.lee, patrick.vandewalle\}@kuleuven.be}
\and
Javier Civera\\
I3A, University of Zaragoza\\
Mar\'{i}a de Luna 1, Zaragoza, Spain\\
{\tt\small jcivera@unizar.es}
}

\begin{document}
\maketitle
\begin{abstract}
We revisit the classical Perspective-Three-Point (P3P) problem, which aims to recover the absolute pose of a calibrated camera from three 2D–3D correspondences.
It has long been known that P3P can be reduced to a quartic polynomial with analytically simple and computationally efficient coefficients.
However, this elegant formulation has been largely overlooked in modern literature. 
Building on the theoretical foundation that traces back to Grunert's work in 1841, we propose a compact algebraic solver that achieves accuracy and runtime comparable to state-of-the-art methods.
Our results show that this classical formulation remains highly competitive when implemented with modern insights, offering an excellent balance between simplicity, efficiency, and accuracy.
\end{abstract}

\section{Introduction}
\label{sec:intro}
Estimating the pose (\ie, position and orientation) of a calibrated camera from image measurements is a foundational problem in computer vision.
It is relevant to a wide variety of applications, including 3D reconstruction \cite{Agarwal}, Structure-from-Motion (SfM) \cite{sfm}, augmented reality (AR) \cite{AR}, visual odometry \cite{vo}, visual localization \cite{localization} and simultaneous localization and mapping (SLAM) \cite{slam}. 
A particularly important and well-studied variant is the Perspective-Three-Point (P3P) problem, where the absolute pose of a calibrated camera is determined from exactly three 2D-to-3D point correspondences (see Fig. \ref{fig:problem}).

The importance of P3P arises not only from its theoretical elegance but also from its practical utility. 
In real-world systems, rapid and reliable pose estimation is critical. 
In AR, for instance, aligning virtual content with the physical world requires accurate and low-latency camera pose estimation.
In SLAM and visual(-inertial) navigation, the camera pose must be tracked accurately from sparse and noisy data. 
In such contexts, the ability to estimate pose from only three correspondences makes P3P a valuable tool, particularly for minimal-solver-based pipelines where efficiency and robustness are crucial.

P3P can also be viewed as a special case of the more general Perspective-$n$-Point (PnP) problem, which estimates camera pose from $n$ 2D-3D correspondences with $n\geq3$.
As the minimal case of PnP, P3P plays an essential role in a RANSAC-based frameworks \cite{fischler1981random}, enabling the robust generation of pose hypotheses from minimal data.
P3P has been a core component in numerous robust estimation pipelines, where its efficiency and stability directly impact the performance of the overall system.

Over the decades, various P3P methods have been proposed, and they are typically classified by the degree and form of the polynomial equations they produce --- most commonly quartic or cubic.
While quartic-based solvers have been traditionally favored for their analytical simplicity, cubic-based solvers have recently attracted increasing interest due to their numerical stability and computational efficiency \cite{Persson_2018_ECCV, Ding_2023_CVPR}.

In this work, we propose a compact quartic-based P3P solver that is numerically stable and computationally efficient. 
Our main contribution is twofold:
\begin{enumerate}
    \item We revisit the formulation of P3P and show that it reduces to a quartic polynomial with coefficients that are remarkably simple to derive and compute. 
    This quartic was previously reported by Smith in 1965 \cite{smith}, while an equivalent form, differing only by a scale factor, can be traced back to Grunert's seminal work in 1841 \cite{grunert1841pothenotische}.
    \item We propose a set of practical implementation strategies, which are shown to significantly enhance the numerical accuracy of the solution.
\end{enumerate}

While many recent P3P solvers derive quartic or cubic polynomials using complex symbolic or geometric tools, our method stands out for its exceptional simplicity. 
It can be derived and understood using only elementary algebraic reasoning, as it does not rely on advanced techniques, such as coordinate reparameterization \cite{kneip_2011_cvpr,masselli} or intermediate coordinate systems \cite{banno,nakano_2019_bmvc}. 
This leads to a solver that is not only efficient and numerically stable but also easy to derive, implement, and verify, making it particularly well-suited for both practical deployment and instructional purposes.


\section{Related Work}
\label{sec:related}
The P3P problem has a long and rich history.
Lagrange discussed this problem as early as from 1773 \cite{historical_survey, lagrange1773solutions, lagrange1795lecons}.
The first complete analytical solution appeared in a paper by Grunert in 1841 \cite{grunert1841pothenotische}.
Grunert formulated the task of determining camera pose from three spatial correspondences and showed that it could yield up to four real solutions. 
His approach relied on geometric constraints and trigonometric relationships, and it became the foundation for early photogrammetric methods (see the survey by Haralick \etal \cite{haralick1991analysis}).
Building on Grunert's ideas, in 1937, Finsterwalder and Scheufele \cite{finsterwalder1937ruckwartseinschneiden} proposed a formulation that effectively reduced the P3P problem to a cubic equation, making their method the first cubic-based P3P solver. 
The practical impact of P3P solvers was amplified by the introduction of RANSAC in 1981 by Fischler and Bolles \cite{fischler1981random}, where they also introduced one of the earliest closed-form quartic-based solvers for P3P.
In 1991, Haralick \etal \cite{haralick1991analysis} offered one of the earliest comprehensive overviews of P3P methods, bringing together and organizing various geometric and algebraic approaches developed over previous decades.

Based on the degree of the polynomial equation they solve, existing P3P solvers can be categorized into two main groups: quartic-based and cubic-based.
Below, we summarize key methods in each group.

\textbf{Quartic-based P3P solvers}:
Quartic-based solvers have been the most common historically.
Gao \etal \cite{gao} applied Wu-Ritt's zero decomposition technique \cite{wu-ritt} to eliminate variables and derive a quartic polynomial whose real roots capture the geometric constraints of the P3P problem. 
Kneip \etal \cite{kneip_2011_cvpr} proposed a parameterization that directly computes the camera pose without estimating 3D points in the camera frame, which reduced the problem to solving a quartic polynomial using two parameters, avoiding the singular value decomposition.
Masselli and Zell \cite{masselli} proposed an alternative prameterization using different geometric constraints.
Ke and Roumeliotis \cite{Ke_2017_CVPR} systematically eliminated the camera position and feature distances, solving directly for orientation by deriving a quartic polynomial in a single rotation parameter.
Banno \cite{banno} introduced an algebraic approach that represents the camera pose parameters as a linear combination of null-space vectors derived from the P3P constraints.
Nakano \cite{nakano_2019_bmvc} improved on Banno’s method by introducing a more efficient coordinate setup, which reduced algebraic complexity and computational cost.
Recently, Wu \etal \cite{Wu_2025_WACV} proposed a conic transformation approach to the P3P problem that maps one of the conics into a canonical parabola, simplifying the intersection computation and eliminating the need for complex arithmetic.

\textbf{Cubic-based P3P solvers}:
A smaller but notable group of methods reduces the P3P problem to solving a cubic polynomial. 
These methods generally aim to reduce the number of candidate solutions and improve numerical stability by simplifying the algebraic structure.
Early examples of this type include the works of Finsterwalder and Scheufele \cite{finsterwalder1937ruckwartseinschneiden} and Grafarend \etal \cite{grafarend1989dreidimensionaler}.
A more recent example is the cubic-based solver proposed by Persson and Nordberg \cite{Persson_2018_ECCV}, where the underlying elliptic equations are solved by diagonalization that requires a single real root of a cubic polynomial. 
In \cite{Ding_2023_CVPR}, Ding \etal revisit the cubic-based approach and systematically characterize all eight possible conic intersection cases. 
By explicitly handling each case, their method improves both numerical stability and computational efficiency over prior work.

\section{Problem Definition}
\label{sec:problem}
\begin{figure}[t]
 \centering
 \includegraphics[width=0.45\textwidth]{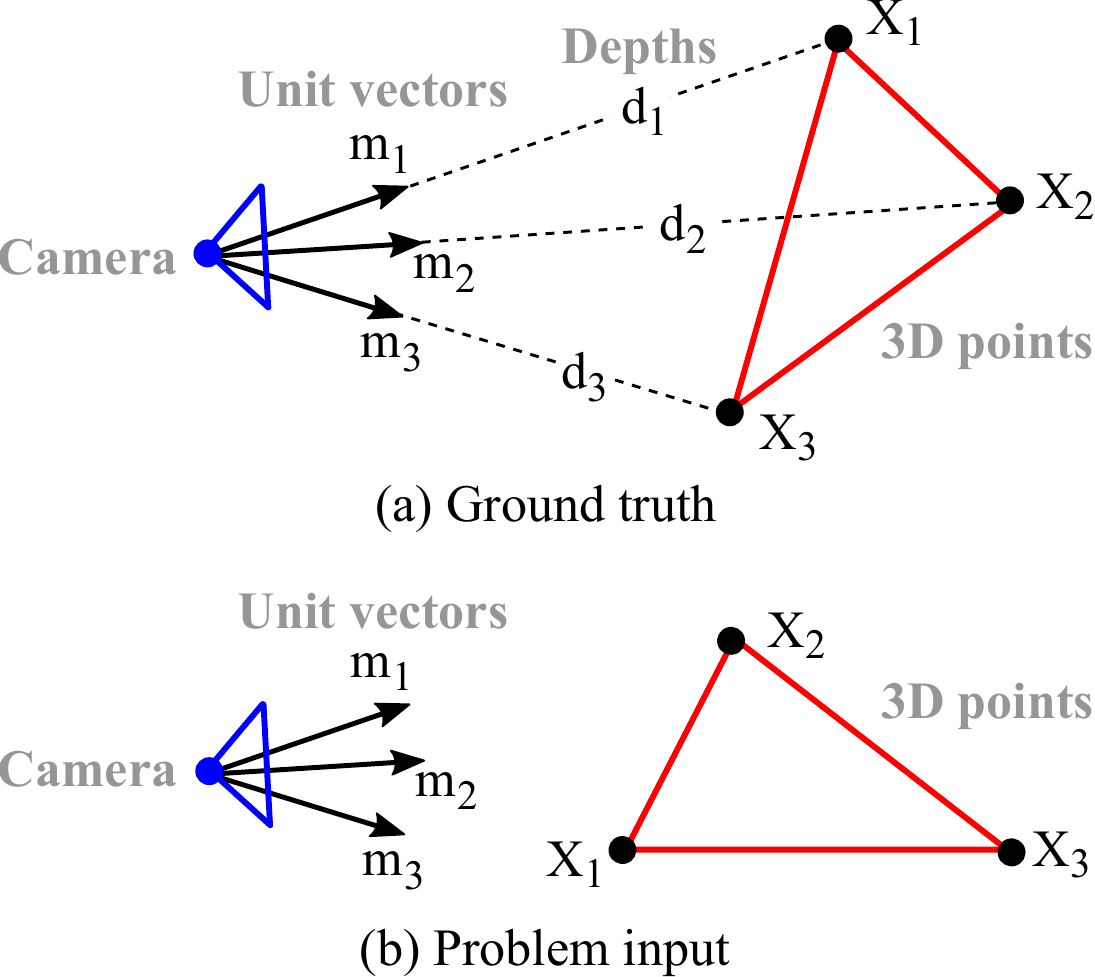}
\caption{An illustration of the P3P problem. \textbf{(a) Ground truth}: The unit bearing vectors from the camera center pass through the corresponding 3D points in space.  \textbf{(b) Problem input}: An unknown rigid-body transformation has been applied to the given 3D points. As a result, the bearing vectors do not pass through the 3D points. Our goal is to find the unknown rigid-body transformation, given these bearing vectors and 3D points as input.
}
\label{fig:problem}
\end{figure}

Consider a calibrated camera located at the origin, observing three 3D points.
Let $\mathbf{m}_1$, $\mathbf{m}_2$ and $\mathbf{m}_3$ be the unit bearing vectors directed toward the corresponding 3D points $\mathbf{X}_1$, $\mathbf{X}_2$ and $\mathbf{X}_3$, as shown in Fig. \ref{fig:problem}.  
Let $\mathbf{R}$ and $\mathbf{t}$ be the unknown rotation matrix and translation vector applied to the 3D points, such that
\begin{align}
    d_1\mathbf{m}_1 &= \mathbf{RX}_1+\mathbf{t},\label{eq:p3p_1}\\
    d_2\mathbf{m}_2 &= \mathbf{RX}_2+\mathbf{t},\\
    d_3\mathbf{m}_3 &= \mathbf{RX}_3+\mathbf{t},\label{eq:p3p_3}
\end{align}
where $d_1$, $d_2$ and $d_3$ are the unknown depths of the points.
Our goal is to find $\mathbf{R}$ and $\mathbf{t}$ (and optionally $\mathbf{d}_1$, $\mathbf{d}_2$ and $\mathbf{d}_3$) that satisfy \eqref{eq:p3p_1}--\eqref{eq:p3p_3}, given the unit vectors $\left(\mathbf{m}_1, \mathbf{m}_2, \mathbf{m_3}\right)$ and the corresponding 3D points $\left(\mathbf{X}_1, \mathbf{X}_2, \mathbf{X}_3\right)$ as input.

\section{Method}
\label{sec:method}
\subsection{Theory revisited}
We define the following variables:
\begin{align}
    m_{12}&=\mathbf{m}_1^\top\mathbf{m}_2, \label{eq:m12}\\
    m_{13}&=\mathbf{m}_1^\top\mathbf{m}_3,\\
    m_{23}&=\mathbf{m}_2^\top\mathbf{m}_3, \label{eq:m23}\\
    s_{12} &= \lVert \mathbf{X}_1-\mathbf{X}_2\rVert^2, \label{eq:s12}\\ 
    s_{13} &= \lVert \mathbf{X}_1-\mathbf{X}_3\rVert^2, \\
    s_{23} &= \lVert \mathbf{X}_2-\mathbf{X}_3\rVert^2. \label{eq:s23}
\end{align}
Note that since $\mathbf{m}_1$, $\mathbf{m}_2$ and $\mathbf{m}_3$ are unit vectors, $m_{12}$,  $m_{13}$ and  $m_{23}$ represent the cosines of the angles between each pair of bearing vectors (see the Fig. \ref{fig:problem}{\color{wacvblue}a}). 
According to the law of cosines, we have
\begin{align}
    d_1^2+d_2^2-2d_1d_2m_{12} &= s_{12}, \label{eq:cosrule1}\\
    d_1^2+d_3^2-2d_1d_3m_{13} &= s_{13},\\
    d_2^2+d_3^2-2d_2d_3m_{23} &= s_{23}.\label{eq:cosrule2}
\end{align}
Here, we introduce the following variables:
\begin{equation}
\label{eq:x_and_y}
    x = \frac{d_1}{d_3}, \quad y= \frac{d_2}{d_3}.
\end{equation}
Substituting $d_1 = xd_3$ and $d_2=yd_3$ into \eqref{eq:cosrule1}--\eqref{eq:cosrule2} and rearranging them, we obtain the following equations:
\begin{align}
    d_3^2 &= \frac{s_{12}}{x^2-2xym_{12}+y^2},\label{eq:d3sq1}\\
    d_3^2 &= \frac{s_{13}}{x^2-2xm_{13}+1},\label{eq:d3sq2}\\
    d_3^2 &= \frac{s_{23}}{y^2-2ym_{23}+1}.\label{eq:d3sq3}
\end{align}
Equating the right-hand side of \eqref{eq:d3sq3} with that of \eqref{eq:d3sq1} and \eqref{eq:d3sq2}, respectively, we get the following two equations:
\begin{equation}
\label{eq:xy1}
    x^2-2m_{12}xy+(1-a)y^2+2am_{23}y-a=0,
\end{equation}
\begin{equation}
\label{eq:xy2}
    x^2-by^2-2m_{13}x+2bm_{23}y+1-b = 0,
\end{equation}
where
\begin{equation}
\label{eq:ab}
a = \frac{s_{12}}{s_{23}}, \quad b = \frac{s_{13}}{s_{23}}.
\end{equation}
Note that \eqref{eq:xy1} and \eqref{eq:xy2} are the same set of equations used in \cite{Ding_2023_CVPR, Wu_2025_WACV} to derive a polynomial equation in $x$.
From here on, however, we take a much simpler approach than these two works:
First, we rearrange \eqref{eq:xy2} as
\begin{equation}
    y^2 = \frac{x^2-2m_{13}x+2bm_{23}y+1-b}{b}.
\end{equation}
Substituting this into \eqref{eq:xy1}, we obtain an equation involving only $x^2$, $x$ and $y$, which we then simplify to isolate $y$ as a rational function of $x$:
\begin{equation}
    y = \frac{(-a+b+1)x^2+2(a-1)m_{13}x-a-b+1}{2b(m_{12}x-m_{23})}.
\end{equation}
Using the definitions in \eqref{eq:ab}, this can also be written as
\begin{equation}
\label{eq:y}
    y=\frac{Ax^2+Bx+C}{2s_{13}(m_{12}x-m_{23})},
\end{equation}
where
\begin{align}
    A &= -s_{12}+s_{23}+s_{13}, \label{eq:A}\\
    B &= 2(s_{12}-s_{23})m_{13},\\
    C &= -s_{12}+s_{23}-s_{13}. \label{eq:C}
\end{align}
Plugging \eqref{eq:y} into \eqref{eq:xy2} yields quartic polynomial in $x$.
Conveniently, it turns out that all of the coefficients of this polynomial contain a common factor, $s_{13}/s_{23}$, which is $b$ from \eqref{eq:ab}.
Factoring this out, we get:
\begin{equation}
\label{eq:quartic}
    c_4x^4+c_3x^3+c_2x^2+c_1x+c_0=0,
\end{equation}
where
\begin{align}
\begin{split}
    c_4={}&-s_{12}^2 + 2s_{12}s_{13} + 2s_{12}s_{23} - s_{13}^2 \\
    &+4s_{13}s_{23}m_{12}^2 - 2s_{13}s_{23} - s_{23}^2,
\end{split} \label{eq:c4}\\
\begin{split}
    c_3={}& 4s_{12}^2m_{13} - 4s_{12}s_{13}m_{12}m_{23} - 4s_{12}s_{13}m_{13}\\
    &- 8s_{12}s_{23}m_{13} + 4s_{13}^2m_{12}m_{23}\\
    &-8s_{13}s_{23}m_{12}^2m_{13} - 4s_{13}s_{23}m_{12}m_{23} \\
    &+ 4s_{13}s_{23}m_{13} + 4s_{23}^2m_{13},
\end{split}\\
\begin{split}
    c_2={}& -4s_{12}^2m_{13}^2 - 2s_{12}^2 + 8s_{12}s_{13}m_{12}m_{13}m_{23} \\
    &+ 4s_{12}s_{13}m_{23}^2 + 8s_{12}s_{23}m_{13}^2 + 4s_{12}s_{23}\\
    &- 4s_{13}^2m_{12}^2 - 4s_{13}^2m_{23}^2 + 2s_{13}^2 \\
    &+ 4s_{13}s_{23}m_{12}^2 + 8s_{13}s_{23}m_{12}m_{13}m_{23}\\
    &- 4s_{23}^2m_{13}^2 - 2s_{23}^2,
\end{split}\\
\begin{split}
    c_1={}& 4s_{12}^2m_{13} - 4s_{12}s_{13}m_{12}m_{23}\\
    &- 8s_{12}s_{13}m_{13}m_{23}^2 + 4s_{12}s_{13}m_{13}\\
    &- 8s_{12}s_{23}m_{13} + 4s_{13}^2m_{12}m_{23}\\
    &- 4s_{13}s_{23}m_{12}m_{23} - 4s_{13}s_{23}m_{13}\\
    &+ 4s_{23}^2m_{13},
\end{split}\\
\begin{split}
    c_0 ={}&-s_{12}^2 + 4s_{12}s_{13}m_{23}^2 - 2s_{12}s_{13}\\
    &+ 2s_{12}s_{23} - s_{13}^2 + 2s_{13}s_{23} - s_{23}^2.
\end{split} \label{eq:c0}
\end{align}
Solving \eqref{eq:quartic} yields up to four real solutions for $x$.
Plugging these $x$ values into \eqref{eq:y} gives the corresponding $y$ values.
By considering only the positive real $x$ and $y$ values, we can find $d_3$ using either \eqref{eq:d3sq1}, \eqref{eq:d3sq2} or \eqref{eq:d3sq3}.
Subsequently, $d_1$ and $d_2$ can be found using \eqref{eq:x_and_y}.
Once all depths are found, we refine them using the Newton-Raphson method on \eqref{eq:cosrule1}--\eqref{eq:cosrule2}.
Afterwards, we align the set of 3D points $\left(d_1\mathbf{m}_1, d_2\mathbf{m}_2, d_3\mathbf{m}_3\right)$ to $\left(\mathbf{X}_1, \mathbf{X}_2, \mathbf{X}_3\right)$ using the method proposed in \cite{Persson_2018_ECCV}:
\begin{align}
    \mathbf{R}&=\begin{bmatrix}
        \mathbf{Y}_1, 
        \mathbf{Y}_2,
        \mathbf{Y}_1\times\mathbf{Y}_2
    \end{bmatrix} \mathbf{X}^{-1}, \label{eq:R}\\   \mathbf{t}&=d_1\mathbf{m}_1-\mathbf{R}\mathbf{X}_1,
\end{align}
where 
\begin{align}
    &\mathbf{Y}_1 =d_1\mathbf{m}_1-d_2\mathbf{m}_2, \quad 
    \mathbf{Y}_2 =d_1\mathbf{m}_1-d_3\mathbf{m}_3,\\
    &\mathbf{X} = \begin{bmatrix}
        \mathbf{X}_1-\mathbf{X}_2, \mathbf{X}_1-\mathbf{X}_3,(\mathbf{X}_1-\mathbf{X}_2)\times(\mathbf{X}_1-\mathbf{X}_3)
    \end{bmatrix}. \label{eq:X}
\end{align}
As a result, we obtain up to four valid solutions for $(\mathbf{R}, \mathbf{t})$, corresponding to the real roots of the quartic.

We note that the quartic formulation \eqref{eq:quartic}--\eqref{eq:c0} was previously derived and used by Smith in 1965 \cite{smith}.
A mathematically equivalent form, differing only by a scale factor, was originally derived by Grunert in 1841 \cite{grunert1841pothenotische}.
Another scaled, yet equivalent, variant was proposed by Fischler and Bolles in 1981 \cite{fischler1981random}.
Table \ref{tab:scaled_variants} compares the quartics employed in these works.
Although all these quartic formulations are mathematically equivalent, we found that ours (identical to Smith’s \cite{smith}) yields the highest numerical precision.
\begin{table}[t]
\centering
\begin{tabular}{cc}
\toprule
{Method} & {Scale factor relative to ours \eqref{eq:quartic}}  \\
\midrule
Grunert (1841) \cite{grunert1841pothenotische} & $\displaystyle\left(s_{13}\right)^{-2}$\\
Smith (1965) \cite{smith}               & $-1$  \\
\makecell[c]{Fischler \& Bolles \\(1981) \cite{fischler1981random}}              & $\displaystyle\left(\frac{s_{12}}{s_{13}s_{23}}\right)^2$  \\
\bottomrule
\end{tabular}
\caption{The relative scale factors between our quartic \eqref{eq:quartic} and those presented in earlier works \cite{grunert1841pothenotische,smith,fischler1981random}.}
\label{tab:scaled_variants}
\end{table}

\subsection{Implementation details}
\label{subsec:implementation}
We found that several subtle implementation choices can have a substantial impact on numerical precision.
When following the procedure described in the previous section, there are several mathematically equivalent paths leading from \eqref{eq:quartic} to the final solution $(\mathbf{R}, \mathbf{t})$.
Empirically, we found that some methods are numerically more precise than the others.
In this section, we report our findings on the numerical impact of some seemingly minor choices. 
Revealing the exact mechanism underlying these heuristics is beyond the scope of this work and is left for future work.
\begin{enumerate}
    \item \textbf{Solving a quartic polynomial:} 
    To solve \eqref{eq:quartic}, we use two different (but mathematically equivalent) versions of Ferrari's method: (1) \textit{Ferrari-Lagrange method} \cite{turnbull}\footnote{It solves a monic quartic polynomial $x^4+ax^3+bx^2+cx+d=0$ using the cubic solvent $y^3-by^2+(ac-4d)y+(4bd-a^2d-c^2)=0$.}, and (2) \textit{Classical Ferrari method}  \cite{cardano}\footnote{It solves a depressed quartic polynomial $u^4+au^2+bu+c=0$ using the cubic solvent $8y^3+20ay^2+(16a^2-8c)y+(4a^3-4ac-b^2)=0$.}.
    A similar approach is proposed in \cite{Wu_2025_WACV}, where Method 1 is used when the leading coefficient has magnitude less than $10^4$, and Method 2 is used when Method 1 could not find any real root or when the leading coefficient has magnitude larger than $10^4$.
    In this work, however, we use a different criterion:
    We use Method 1 when $\displaystyle\left|\frac{c_3}{c_4}\right|>10$; otherwise, we use Method 2.
    In both Method 1 and Method 2, we discard a root if its computation involves a negative argument under a square root, as in \cite{Wu_2025_WACV}, to ensure that all intermediate computations remain real-valued.
    For the detailed description of the two methods, see the supplementary material of \cite{Wu_2025_WACV}.

    \item \textbf{Reindexing the points:} 
    We found that indexing the points in a certain order improves the numerical accuracy of the results.
    In fact, a similar trick has been already known among some of the researchers (see the open-source code of \cite{nakano_2019_bmvc, Ding_2023_CVPR} for example).
    Specifically, we reindex the points, if necessary, to ensure that 
    \begin{equation}
    \label{eq:index}
        m_{13}\leq m_{12} \leq m_{23}.
    \end{equation}
    Note that this is different from \cite{nakano_2019_bmvc,Ding_2023_CVPR} where the indexing is based on the relative magnitudes of $s_{12}$, $s_{13}$ and $s_{23}$.
    
    \item \textbf{Finding $\mathbf{d}_\mathbf{3}$:}
    Once $x$ and $y$ values are found, we can choose to use either \eqref{eq:d3sq1}, \eqref{eq:d3sq2} or \eqref{eq:d3sq3} to compute $d_3$.
    Theoretically, the choice of equation should not matter.
    In practice, however, we found that using \eqref{eq:d3sq3} achieves higher numerical accuracy than using the other two.
\end{enumerate}
Our method is summarized in Alg. \ref{algorithm}.

\begin{algorithm}
\caption{Proposed P3P method}
\label{algorithm}
\textbf{Input:} 3D points, $\mathbf{X}_i$, and the corresponding unit bearing vectors, $\mathbf{m}_i$, $i=1,2,3$. \\
\textbf{Output:} Poses $\{\mathbf{R}_j, \mathbf{t}_j\}_{j=1,\dots,N}$, where $N \leq 4.$
\begin{algorithmic}[1]
\State Compute $m_{12}, m_{13}, m_{23}$ using \eqref{eq:m12}--\eqref{eq:m23}.
\State Reindex, if necessary, the points and bearing vectors so that \eqref{eq:index} is satisfied.
\State Compute $s_{12}, s_{13}, s_{23}$ using \eqref{eq:s12}--\eqref{eq:s23}.
\State Compute $c_4, c_3, c_2, c_1, c_0$ using \eqref{eq:c4}--\eqref{eq:c0}.
\If{$\displaystyle\left|\frac{c_3}{c_4}\right|>10$}
\State Solve \eqref{eq:quartic} using \textit{Ferrari-Lagrange method} \cite{turnbull}.
\Else
\State Solve \eqref{eq:quartic} using \textit{Classical Ferrari method} \cite{cardano}. 
\EndIf
\For{every positive real root $x$}
    \State Compute $y$ using \eqref{eq:y}--\eqref{eq:C}.
    \If{$y>0$}
    \State Compute $d_3$ using \eqref{eq:d3sq3}.
    \State Compute $d_1$ and $d_2$ using \eqref{eq:x_and_y}.
    \State Refine $d_1, d_2, d_3$ using Newton-Raphson method on \eqref{eq:cosrule1}--\eqref{eq:cosrule2}.
    \State Compute $\mathbf{R}$ and $\mathbf{t}$ using \eqref{eq:R}--\eqref{eq:X}.
    \EndIf
\EndFor
\end{algorithmic}
\end{algorithm}

\section{Results}
\label{sec:results}
In this section, we compare our method against several state-of-the-art solvers: namely, the ones by Wu \etal \cite{Wu_2025_WACV}, Ding \etal \cite{Ding_2023_CVPR}\footnote{We test both the original version, \texttt{Ding\_old}, published in November 2023, and the updated version, \texttt{Ding\_new}, published in April 2024. For both versions, we use the `Adjoint' method, as recommended by the authors.}, Persson and Nordberg \cite{Persson_2018_ECCV}, Nakano \cite{nakano_2019_bmvc}\footnote{The author of \cite{nakano_2019_bmvc} published only the MATLAB code. To ensure a fair comparison, we use the C++ implementation provided by the authors of \cite{Ding_2023_CVPR} at \url{https://github.com/yaqding/P3P} (as in \cite{Wu_2025_WACV}).}, Ke and Roumeliotis \cite{Ke_2017_CVPR}, and Kneip \etal \cite{kneip_2011_cvpr}.
All methods are implemented in C++ and run on a laptop CPU with i7-7700HQ 2.8GHz.
Our evaluation strategy is aligned with the approaches used in \cite{Persson_2018_ECCV,Ding_2023_CVPR,Wu_2025_WACV}.
First, the synthetic dataset is generated as follows:
\begin{enumerate}
    \item Generate three random image points with coordinates $(u_i, v_i)$ such that $u_i, v_i \sim \mathcal{U}(-1,1)$ for $i=1,2,3$. 
    \item Generate the unit bearing vectors, $\mathbf{m}_i$:
    \begin{equation}
        \mathbf{m}_i=\frac{[u_i,v_i,1]^\top}{\sqrt{u_i^2+v_i^2+1}}.
    \end{equation}
    \item Generate the corresponding 3D points $\mathbf{X}_i$ with a random depth $d_i\sim\mathcal{U}(0.1, 10)$:
    \begin{equation}
    \label{eq:generate_depth}
        \mathbf{X}_i = d_i\mathbf{m}_i.
    \end{equation}
    \item Generate a random ground-truth rotation matrix $\mathbf{R}_\text{gt}$ and a translation vector of length $1$, $\mathbf{t}_\text{gt}$.
    \item Apply this random rigid body transformation to the points:
    \begin{equation}
        \mathbf{X}_i \gets \mathbf{R}_\text{gt}^\top(\mathbf{X}_i - \mathbf{t}_\text{gt}).
    \end{equation}
\end{enumerate}

Next, we evaluate each method on the same dataset and report the pose error used in \cite{Ding_2023_CVPR, Wu_2025_WACV}:
\begin{equation}
    \xi = \lVert\mathbf{R}_\text{gt}-\mathbf{R}_\text{est}\rVert_{L1}+\lVert\mathbf{t}_\text{gt}-\mathbf{t}_\text{est}\rVert_{L1}.
\end{equation}
Based on this error metric, we determine how many solutions fall into each of the following categories (as in \cite{Ding_2023_CVPR, Wu_2025_WACV}):
\begin{enumerate}
    \item \textit{Valid solutions}: All solutions $(\mathbf{R}, \mathbf{t})$ returned by the solver.
    \item \textit{Unique solutions}: Solutions that have a valid rotation matrix\footnote{We consider a rotation matrix to be valid if it meets the following conditions: (1) $|\det{(\mathbf{R})-1}|<10^{-6}$, (2) $\lVert\mathbf{R}^\top\mathbf{R}-\mathbf{I}\rVert_{L1}<10^{-6}$, and (3) $|1-\text{Quaternion norm of } \mathbf{R}|<10^{-5}$.} and a reprojection error in the normalized image coordinates below $10^{-4}$, after removing the duplicates.
    \item \textit{Duplicates}: When two solutions $(\mathbf{R}_1, \mathbf{t}_1)$ and $(\mathbf{R}_2, \mathbf{t}_2)$ satisfy $\lVert\mathbf{R}_1-\mathbf{R}_2\rVert_{L1}+\lVert\mathbf{t}_1-\mathbf{t}_2\rVert_{L1}<10^{-5}$, we say that we have a duplicate. 
    \item \textit{Good solutions}: The number of simulated problems where at least one unique solution is found.
    \item \textit{No solution}: The number of simulated problems where no unique solution is found. 
    \item \textit{Ground truth}: The number of simulated problems where at least one solution satisfies $\xi<10^{-6}.$
    \item \textit{Incorrect}: The number of valid solutions (including the duplicates) that do not meet the criteria for unique solutions.
\end{enumerate}

The main results are presented in Tab. \ref{tab:method_comparison}.
It can be seen that Ding\_new \cite{Ding_2023_CVPR} is the most robust method.
In terms of the number of ground-truth solutions obtained, our method is the second best:
The difference between ours and Ding\_new \cite{Ding_2023_CVPR} is only $8$ out of $10^8$ problems, amounting to a statistically negligible difference of 0.000008\% of the total problems.
In terms of the number of good solutions, the difference is only 4 out of $10^8$, which amounts to 0.000004\%.
We also found that the method by Wu \etal \cite{Wu_2025_WACV} performs slightly worse than the authors reported: 
In \cite{Wu_2025_WACV}, it achieves the ground truth in 99.99997\% of cases, while in our evaluation, it does so in 99.99989\% of cases.
We speculate that two potential causes may account for this discrepancy.
First, we use a sample size that is 10 times larger than that used in \cite{Wu_2025_WACV}, so there is a lower risk of sampling bias.
Second, the authors of \cite{Wu_2025_WACV} might have mistakenly simulated the depths $d_i$ in \eqref{eq:generate_depth} using a uniform distribution within the interval $[0.1, 100]$, instead of $[0.1, 10]$ as stated in their paper\footnote{We suspect this for two main reasons: First, the authors' public source code uses the distribution $\mathcal{U}(0.1,100)$ for the depths, and the file containing this line is identical to one uploaded by the authors of \cite{Ding_2023_CVPR} where the same distribution is used. Second, our experiment confirmed that using the distribution $\mathcal{U}(0.1,100)$ for the depths improves the results of their method, producing numbers very close to those reported in \cite{Wu_2025_WACV}.}.

\begin{table*}[ht]
\small
\centering
\begin{tabular}{ccccccccc}
\toprule
\makecell{Method} 
& \makecell{Ours} 
& \makecell{Wu \\ \cite{Wu_2025_WACV}} 
& \makecell{Ding\_new \\ \cite{Ding_2023_CVPR}} 
& \makecell{Ding\_old \\ \cite{Ding_2023_CVPR}} 
& \makecell{Persson \\ \cite{Persson_2018_ECCV}} 
& \makecell{Ke \\ \cite{Ke_2017_CVPR}} 
& \makecell{Kneip \\ \cite{kneip_2011_cvpr}} 
& \makecell{Nakano \\ \cite{nakano_2019_bmvc}} \\
\midrule
Valid     & 168853631& 168853618& 168853640 & 168853633 & 168853625 & 174550801 & 242373408 & 196891118 \\
Unique    & 168853586 & 168853563 & 168853619 & 168853604 & 168853528 & 168866306 & 168876463 & 168844644 \\
Duplicates ($\downarrow$)          & 22 & 18 & 17 & \textbf{16} & 20 & 1893830 & 29666 & 108 \\
Good ($\uparrow$)     & 99999995 & 99999984 & \textbf{99999999} & 99999996 & 99999969 & 99997580 & 99997085 & 99994088 \\
No solution ($\downarrow$)        & 5& 16&  \textbf{1}&  4&  31&  2420&  2915& 5912 \\
Ground truth ($\uparrow$)       & 99999979& 99999890& \textbf{99999987} & 99999933 & 99999777 & 99989030 & 99957466 & 99993317 \\
Incorrect ($\downarrow$) & 23 & 37 & \textbf{4} & 11 & 77 & 3790665 & 73472739 & 28046366 \\
\bottomrule
\end{tabular}
\caption{Accuracy comparison of the eight solvers on $10^8$ simulated problems. 
The symbol $\uparrow$ indicates that the higher the better. 
The symbol $\downarrow$ indicates that the lower the better. 
The best results are highlighted in bold.}
\label{tab:method_comparison}
\end{table*}
\begin{table*}[ht]
\centering
\small
\begin{tabular}{ccccccccc}
\toprule
\makecell{Time (ns)} 
& \makecell{Ours} 
& \makecell{Wu \cite{Wu_2025_WACV}} 
& \makecell{Ding\_new \cite{Ding_2023_CVPR}} 
& \makecell{Ding\_old  \cite{Ding_2023_CVPR}} 
& \makecell{Persson \cite{Persson_2018_ECCV}} 
& \makecell{Ke \cite{Ke_2017_CVPR}} 
& \makecell{Kneip \cite{kneip_2011_cvpr}} 
& \makecell{Nakano \cite{nakano_2019_bmvc}} \\
\midrule
Mean     & 254.7 & \textbf{248.1} & 257.2 & 267.8  & 305.9 & 515.1& 826.5 & 1071.7 \\
Median   & 256.1  & \textbf{248.0} & 256.6 & 270.5 & 304.3 & 514.8 & 831.9 & 1075.7  \\
Minimum  & 237.2 & \textbf{228.5}  & 236.5 & 247.7 & 286.6 & 467.1 & 747.2  & 954.9 \\
Maximum  & \textbf{280.2} & 290.9 & 292.4 & 330.1 & 345.1 & 608.7 & 937.8 & 1434.8  \\
\bottomrule
\end{tabular}
\caption{Time comparison of the eight solvers on $10^8$ simulated problems: For each individual problem, we repeat the test 10 times and average the measurements. Then, we report the mean, median, minimum and maximum of these $10^8$ average results.}
\label{tab:time_comparison}
\end{table*}
\begin{table*}[ht]
\footnotesize
\centering
\begin{tabular}{lcccccccc}
\toprule
\makecell{Method} 
& \makecell{Ours \\(Baseline)} 
& \makecell{Using only \\ \textit{Ferrari-Lagrange} \\ \textit{method} \cite{turnbull}} 
& \makecell{Using only \\ \textit{Classical Ferrari} \\ \textit{method} \cite{cardano}} 
& \makecell{Switching \\ criterion \\ of \cite{Wu_2025_WACV} } 
& \makecell{Without \\ reindexing} 
& \makecell{Reindexing  \\ of \cite{nakano_2019_bmvc,Ding_2023_CVPR}} 
& \makecell{Using \eqref{eq:d3sq1} \\ instead of \\ \eqref{eq:d3sq3}} 
& \makecell{Using \eqref{eq:d3sq2} \\ instead of \\ \eqref{eq:d3sq3}}  \\
\midrule
Valid     & 168853631& 168853016&  168853148& 168853551 & 168853457 & 168853622 &  168853631& 168853631  \\
Unique    & 168853586 & 168852752 & 168852972 & 168853311 & 168853340 & 168853578 &  168853574& 168853574  \\
Duplicates ($\downarrow$)         & 22 & 84 & 50 & 82 & 41 & 22 &  \textbf{21}& \textbf{21}  \\
Good ($\uparrow$)     & \textbf{99999995} & 99999700 & 99999867 & 99999974 & 99999951 & 99999994 &  99999992& 99999992  \\
No solution  ($\downarrow$)       & \textbf{5}& 300& 133 & 26 & 49 & 6 &  8& 8  \\
Ground truth ($\uparrow$)       & \textbf{99999979}&99999600 & 99999645 & 99999891 &  99999836& 99999964 &  99999948& 99999946  \\
Incorrect ($\downarrow$) & 23 & 180 & 126 & 158 & 76 & \textbf{22} &  36& 36  \\
\bottomrule
\end{tabular}
\caption{Ablation study on $10^8$ simulated problems:
We evaluate the impact of the heuristics described in Section \ref{subsec:implementation}.}
\label{tab:ablation}
\end{table*}
Table \ref{tab:mean_median_max} compares the mean, median and maximum errors of the ground-truth solutions produced by each method.
Looking at the mean errors, Ding\_new \cite{Ding_2023_CVPR} once again demonstrates the highest numerical stability, with our method ranking a close second.
With an average numerical difference of just $1.8\times10^{-13}$, the two methods exhibit a comparable level of numerical stability.

Table \ref{tab:time_comparison} compares the running times of each method.
Overall, the two fastest methods are the one by Wu \etal \cite{Wu_2025_WACV} and ours.
Looking at the mean times, the method by Wu \etal \cite{Wu_2025_WACV} is faster than ours, but only by 6.6 nanoseconds on average, a difference that is practically negligible.

Finally, in Tab. \ref{tab:ablation}, we conduct an ablation study to evaluate the impact of the heuristics described in Section \ref{subsec:implementation}.
The result illustrates that even seemingly minor details can have a non-negligible impact in practice.

\section{Discussion}
\label{sec:discussion}
\subsection{Practical significance}
By reducing the P3P problem to solving a single quartic polynomial with straightforward coefficients, we avoid the complexity often seen in other approaches. 
Despite its simplicity, our method achieves accuracy and runtime on par with state-of-the-art solvers. 
In many cases, it even outperforms more elaborate methods in terms of stability and robustness. 
This demonstrates that, in some occasions, well-designed and carefully implemented simple formulations match or even outperform more sophisticated alternatives. 
Similar findings have been reported in other perception tasks, e.g.,~\cite{musgrave2020metric,vizzo2023kiss}.

Although our method does not outperform Ding\_new \cite{Ding_2023_CVPR}, the numerical difference is so small that the method that performs most similarly to Ding\_new in terms of number of ground-truth solutions returned is ours and not their original CVPR version (Ding\_old).
Since P3P is almost always used within RANSAC \cite{fischler1981random} (or its variants), the actual practical difference between ours and Ding\_new is essentially negligible.

\subsection{Limitations and future work}
Unfortunately, the heuristics proposed in Sec.~\ref{subsec:implementation} currently lack a clear theoretical justification. 
The precise numerical mechanisms that explain their effectiveness remain unclear. 
In particular, the quartic root-finding step is the most numerically sensitive component of the pipeline. 
Our adaptive selection between two Ferrari-based solvers alleviates this issue, but certain edge cases persist where existing methods, such as Ding\_new \cite{Ding_2023_CVPR}, exhibit slightly better numerical stability.
Further exploration of robust root-finding strategies or methods exploiting the specific structure of our quartic could potentially improve both reliability and efficiency.
We leave this for future work.

\section{Conclusion}
\label{sec:conclusion}
In this work, we revisited the classical P3P problem and demonstrated that it can be solved efficiently by finding the roots of a quartic polynomial with compact, analytically simple coefficients.
This quartic formulation, originally presented by Smith in 1965 \cite{smith}, is mathematically equivalent to those of Grunert (1841) \cite{grunert1841pothenotische} and Fischler and Bolles (1981) \cite{fischler1981random}, differing only by scale factors.
Several implementation strategies were introduced to enhance numerical stability, leading to accuracy and runtime performance comparable to state-of-the-art methods.

Beyond its empirical results, the algebraic simplicity of our formulation makes it well suited for teaching geometric vision concepts, deploying lightweight systems on edge devices, and serving as a clear baseline for future research on minimal solvers.
Our work also illustrates that classical formulations, when revisited with modern insight, can rival or even surpass many recent, more elaborate methods.
Our code is publicly available at 
{\url{https://github.com/sunghoon031/P3P}}.

\begin{table}[ht]
\centering
\small
\begin{tabular}{cccc}
\toprule
{Method} & {Mean} & {Median} & {Maximum} \\
\midrule
Ours               & 1.514e-12 & 3.558e-14 & 9.662e-7  \\
Wu \cite{Wu_2025_WACV} &  6.918e-12 & 3.266e-14 & 9.666e-7\\
Ding\_new \cite{Ding_2023_CVPR} &  \textbf{1.335e-12} & 2.266e-14 & \textbf{8.306e-7}\\
Ding\_old \cite{Ding_2023_CVPR} &  4.647e-12 & 3.010e-14 & 9.925e-7 \\
Persson \cite{Persson_2018_ECCV} & 1.082e-11 & 4.216e-14 & 9.960e-7  \\
Ke \cite{Ke_2017_CVPR}          & 9.195e-11 & \textbf{1.835e-14} & 1.000e-6\\
Kneip \cite{kneip_2011_cvpr}    & 2.756e-10 & 4.317e-14 & 1.000e-6\\
Nakano \cite{nakano_2019_bmvc}  & 1.694e-12 & 2.526e-14 & 9.589e-7\\
\bottomrule
\end{tabular}
\caption{Comparison of the mean, median and maximum errors of the ground-truth solutions returned by each method (given in the second row from the bottom of Tab. \ref{tab:method_comparison}).}
\label{tab:mean_median_max}
\end{table}


\end{document}